\documentclass[11pt]{article}

\usepackage{acl}

\usepackage{times}
\usepackage{latexsym}

\usepackage[T1]{fontenc}

\usepackage[utf8]{inputenc}

\usepackage{microtype}

\usepackage{inconsolata}

\usepackage{graphicx}
\usepackage{booktabs}
\usepackage{amsmath}
\usepackage{makecell}
\usepackage{enumitem}
\usepackage{pifont}
\usepackage{multirow}

\newcommand{\cmark}{\textcolor{green}{\ding{51}}}
\newcommand{\xmark}{\textcolor{red}{\ding{55}}}

\title{Market-Bench: Benchmarking Large Language Models on Economic and Trade Competition}

\author{
  \textbf{Yushuo Zheng\textsuperscript{1,2}},
  \textbf{Huiyu Duan\textsuperscript{1,*}},
  \textbf{Zicheng Zhang\textsuperscript{1,2}},
  \textbf{Yucheng Zhu\textsuperscript{1}},
  \textbf{Xiongkuo Min\textsuperscript{1,*}},
  \textbf{Guangtao Zhai\textsuperscript{1,2,*}}
\\
\\
  \textsuperscript{1}Shanghai Jiao Tong University,
  \textsuperscript{2}Shanghai Artificial Intelligence Laboratory
\\
  \small{
    \{yushuozheng, huiyuduan, zzc1998, zyc420, minxiongkuo, zhaiguangtao\}@sjtu.edu.cn
  } \\
  \small{\textsuperscript{*}Corresponding author.} \\
    \url{https://github.com/aiben-ch/Market-Bench}
}

\begin{document}

\maketitle

\begin{abstract}
The ability of large language models (LLMs) to manage and acquire economic resources remains unclear. In this paper, we introduce \textbf{Market-Bench}, a comprehensive benchmark that evaluates the capabilities of LLMs in economically-relevant tasks through economic and trade competition.
Specifically, we construct a configurable multi-agent supply chain economic model where LLMs act as retailer agents responsible for procuring and retailing merchandise.
In the \textbf{procurement} stage, LLMs bid for limited inventory in budget-constrained auctions. In the \textbf{retail} stage, LLMs set retail prices, generate marketing slogans, and provide them to buyers through a role-based attention mechanism for purchase.
Market-Bench logs complete trajectories of bids, prices, slogans, sales, and balance-sheet states, enabling automatic evaluation with economic, operational, and semantic metrics. Benchmarking on 20 open- and closed-source LLM agents reveals significant performance disparities and winner-take-most phenomenon, \textit{i.e.}, only a small subset of LLM retailers can consistently achieve capital appreciation, while many hover around the break-even point despite similar semantic matching scores. Market-Bench provides a reproducible testbed for studying how LLMs interact in competitive markets.
\end{abstract}

\section{Introduction}

\begin{figure}[t]
    \centering
    \includegraphics[width=\columnwidth]{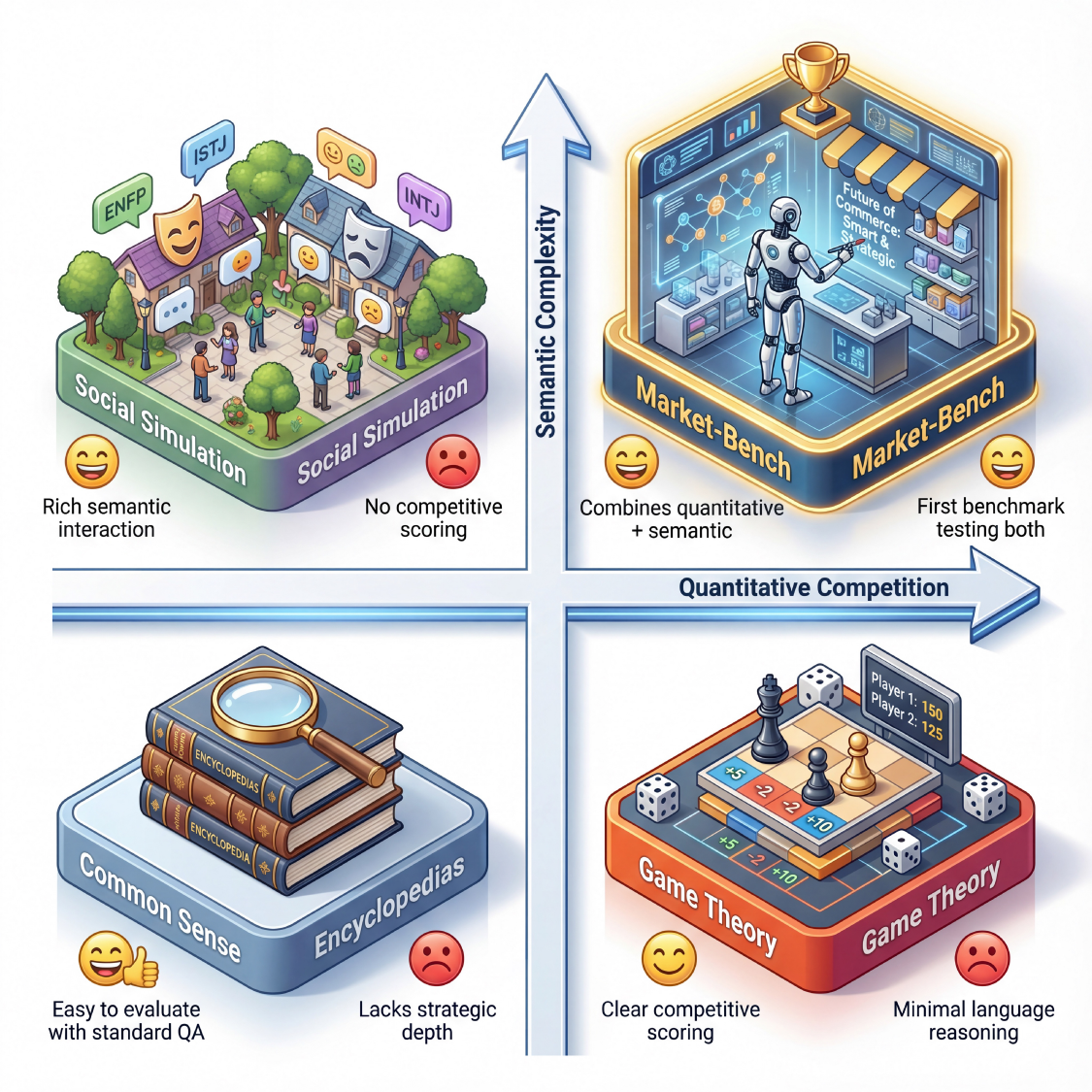}
\caption{Existing LLM benchmarks focus on either semantic complexity or quantitative competition, but rarely both simultaneously under economic scarcity. Thus, we propose Market-Bench, coupling marketing slogans and operations to jointly evaluate mathematical optimization and language comprehension.}
\label{fig:mercantilbench_teaser}
\end{figure}

The advancement of large language models (LLMs) has driven the application of AI in retail from passive analytics to active supply chain management and personalized marketing~\cite{mckinsey2023retail, statista2024ai}, requiring both the \textit{computational precision} to manage scarce resources and the \textit{semantic flexibility} to construct persuasive narratives. Marketing language is a decisive economic variable that stimulates purchase intentions by arousing consumers' latent motivations~\cite{keller2003strategic, kohli2007slogans}, and recent work has begun to formalize the aesthetic evaluation of commercial imagery~\citep{ji2026a3advertisingaestheticassessment}. However, current LLM benchmarks largely evaluate these capabilities in isolation. As shown in Figure~\ref{fig:mercantilbench_teaser}, existing frameworks typically focus either on abstract strategic rationality (``Homo Economicus'') or open-ended social role-play (``Homo Loquens''), rarely demanding both simultaneously. Social simulations~\cite{park2023generative, agentsociety2025} model free-form dialogue but impose no hard economic constraints such as scarcity or bankruptcy, while economic simulators~\cite{dwarakanath2024abides, alympics2025} test rigorous quantitative logic but abstract away linguistic complexity. Although LLM evaluation has expanded rapidly across visual quality assessment~\citep{duan2025finevq, duan2022confusing, duan2023attentive, zhang2026versatile, zhang2025teaching, jin2025rgcvqa, wang2025learning}, multimodal reasoning~\citep{zhang2025large, zhang2026aibench, yang2025odibench, liu2024fbench, zheng2026learningwanderimprovingglobal, zheng2025geoxbench}, and domain-specific fields such as medicine~\citep{ji2025medomni, ji2025evaluating, ji2025assessing, ji2024application, wang2026dental}, no existing benchmark unifies multi-turn buyer--seller interactions across both numeric and semantic dimensions under strict constraintsthat is, a testbed for \textit{dual-process reasoning} within a closed-loop economy remains absent.

This raises a fundamental question: \textit{Can contemporary LLMs effectively perform quantitative reasoning and semantic adaptation to survive in markets characterized by scarcity and competition?}
The market domain is uniquely challenging because optimal pricing and procurement fail without persuasive marketing, while strong marketing fails if budgeting or inventory is mismanaged,a strict co-dependency absent from benchmarks that test either capability alone. Moreover, agents must infer hidden buyer preferences from limited numerical signals under multi-turn competitive dynamics, where errors compound into bankruptcy rather than minor accuracy drops.

We introduce \textbf{Market-Bench}, a configurable multi-agent supply-chain economy formulated as a Partially Observable Markov Game with financial and physical constraints. At each step, agents receive structured text observations,private funds and inventory, auction parameters, and public market history,and output strictly formatted JSON combining discrete mathematical actions (procurement bids, retail prices) with a free-form marketing slogan. Crucially, we propose \textbf{Persona-Gated Attention} (PGA), a mechanism grounded in consumer ``consideration set'' theory~\cite{keller2003strategic} that operationalizes free-form language as a computable, economically consequential variable. Unlike subjective LLM-as-judge heuristics, PGA provides a strict mathematical gatekeeper: it dynamically computes semantic alignment between generated slogans and hidden buyer personas to determine market visibility, directly linking language generation to financial outcomes.

Our contributions are as follows:
\begin{itemize}
    \item We define \textbf{Market-Bench}, a closed-loop economic environment that enforces hard scarcity (finite funds, bankruptcy) and competitive exclusion.
    \item We propose \textbf{Persona-Gated Attention}, a mechanism that operationalizes the economic value of language, requiring agents to optimize semantic similarity to latent buyer personas to secure market access.
    \item We provide a reproducible benchmark implementation and report results across 20 LLM agents using automatically computed economic, supply-chain, and semantic metrics.
\end{itemize}

\begin{figure*}[t]
    \includegraphics[width=\textwidth]{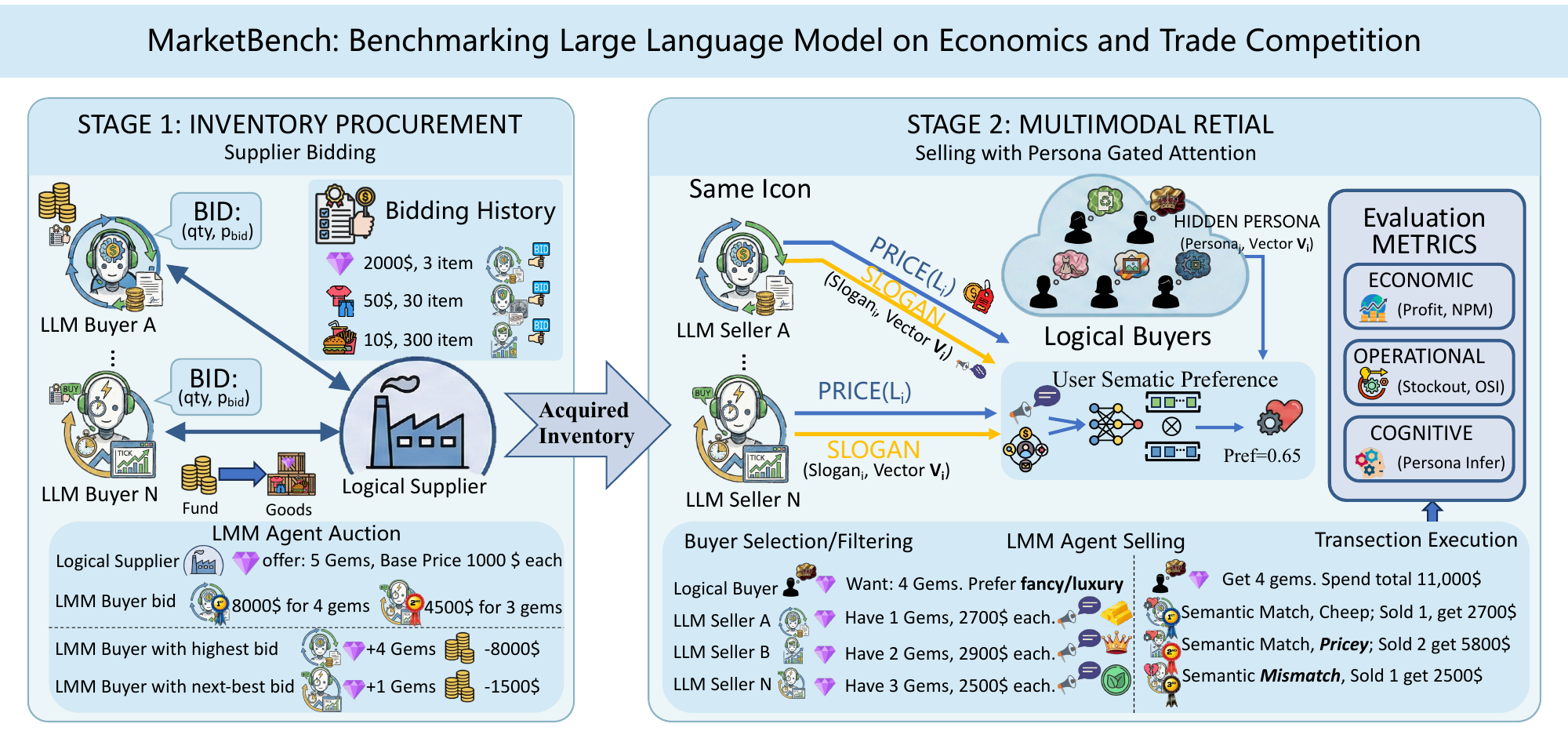}
    \caption{Overview of the Market-Bench environment. Agents operate in a competitive supply chain economy, making procurement, pricing, and marketing decisions under scarcity conditions. The Persona-Gated Attention mechanism requires agents to align their marketing slogans with latent buyer personas to convert demand into sales. Performance is evaluated across economic, supply chain, and cognitive metrics.}
    \label{fig:mercantilbench_overview}
\end{figure*}
\section{Related Work}

\begin{table}[t]
\centering
\small
\setlength{\tabcolsep}{3pt}
\renewcommand{\arraystretch}{1.05}
\resizebox{\columnwidth}{!}{%
\begin{tabular}{l c c c c c}
\toprule
\textbf{Work} & 
\textbf{\makecell{Competitive\\Interaction}} & 
\textbf{\makecell{Hard\\Scarcity}} & 
\textbf{\makecell{Closed-Loop\\Economy}} & 
\textbf{\makecell{Free-Form\\Language}} & 
\textbf{\makecell{Text$\to$\\Payoff}} \\
\midrule
\multicolumn{6}{l}{\textit{Agent Societies}} \\
\cite{park2023generative} & \xmark & \xmark & \xmark & \cmark & \xmark \\
\cite{sotopia2023} & \xmark & \xmark & \xmark & \cmark & \xmark \\
\cite{agentsociety2025} & \xmark & \xmark & \xmark & \cmark & \xmark \\
\midrule
\multicolumn{6}{l}{\textit{Strategic Games}} \\
\cite{multiagentbench2025} & \cmark & \xmark & \xmark & \cmark & \xmark \\
\cite{llmspark2025} & \cmark & \xmark & \xmark & \cmark & \xmark \\
\cite{alympics2025} & \cmark & \cmark & \xmark & \cmark & \xmark \\
\cite{negotiation2025} & \cmark & \xmark & \xmark & \cmark & \cmark \\
\midrule
\multicolumn{6}{l}{\textit{Economic Simulators}} \\
\cite{dwarakanath2024abides} & \xmark & \xmark & \cmark & \xmark & \xmark \\
\midrule
\textbf{Market-Bench} & \cmark & \cmark & \cmark & \cmark & \cmark \\
\bottomrule
\end{tabular}%
}
\captionsetup{justification=raggedright, singlelinecheck=false}
\caption{Comparison of representative LLM-agent benchmarks and economic simulators. \textbf{Market-Bench} is the only benchmark providing both a closed-loop multi-agent market under hard scarcity and free-form language that directly affects payoff via Persona-Gated Attention.}
\label{tab:benchmark-comparison}
\end{table}

\subsection{LLM-Based Agent Societies}
Generative agents have shifted evaluation from static QA to dynamic sandboxes. \citet{park2023generative} showed LLMs can simulate believable social behavior, and recent frameworks like \textit{AgentSociety}~\cite{agentsociety2025} and \textit{MultiAgentBench}~\cite{multiagentbench2025} have scaled these simulations to urban environments and diverse tasks. However, these benchmarks primarily assess social believability or cooperative completion, rarely imposing the ``scarcity conditions'' (finite funds, bankruptcy) characteristic of real economies. \textbf{Market-Bench} bridges this gap by enforcing strict financial survival ($Funds_i < 0 \Rightarrow Bankruptcy$) within a competitive oligopoly.

\subsection{Strategic Reasoning and Game Theory}
A distinct line of research evaluates LLMs as ``Homo Economicus'', testing their rationality in classic game-theoretic setups. \citet{llmspark2025} and \citet{alympics2025} subject agents to matrix games like Prisoner's Dilemma~\cite{axelrod1984evolution} or Trust Game~\cite{berg1995trust} to measure cooperation, deception, and Nash equilibrium convergence. Similarly, \citet{negotiation2025} explores bilateral bargaining, focusing on whether LLMs can reach Pareto-optimal deals. \citet{zheng2025lmfightarenabenchmarking} further demonstrate the effectiveness of competitive game settings for revealing capability gaps among multimodal models. While these benchmarks test strategic logic, they often abstract away the linguistic complexity of commerce, reducing interaction to selecting a move (Cooper/Defect) or outputting a number. \textbf{Market-Bench} addresses this limitation by integrating free-form language as a first-class action, where agents must generate persuasive slogans to unlock market access through our novel \textbf{Persona-Gated Attention} mechanism.

\subsection{Supply Chain and Business Optimization}
LLMs are increasingly deployed as decision-support tools for supply chain management~\cite{simchilevi2025supply, dwarakanath2024abides}. \citet{dwarakanath2024abides} introduce \textit{ABIDES-Economist} for macroeconomic simulation, and Anthropic's Project Vend~\cite{projectvend2025} deployed Claude to autonomously run a retail shop in a single-agent setting. These systems generally position the LLM as a backend optimizer rather than an autonomous agent surviving competitive pressure. \textbf{Market-Bench} unifies procurement, pricing, and marketing into a single scalable loop under active multi-agent competition.

\section{Market-Bench}
\label{sec:mercantilbench}
Market-Bench is a competitive supply-chain economy designed to test \emph{dual-process} agent behavior in a
single closed loop: agents must (i) optimize numeric decisions under budget and inventory constraints, and
(ii) generate natural language marketing that determines which consumers can even \emph{see} them.
Crucially, this mechanism turns the simulation into both an evaluation environment and a \emph{generative
dataset}: each episode yields structured trajectories of observations, numeric actions, text actions, and
economic outcomes, enabling fine-grained analysis beyond static QA-style benchmarks.
Figure~\ref{fig:mercantilbench_overview} summarizes the lifecycle.

\subsection{Economic Setting and State Variables}
We model a two-sided market with an upstream supplier and downstream retailers.
Let 
\begin{equation}
\mathcal{A}=\{A_1,\dots,A_m\}   
\end{equation}

be the set of retailer agents, and let $\mathcal{X}$ denote the set of items.
At step $t$, the environment state includes each agent's funds $Funds_i(t)$ and inventory $Inv_{i,x}(t)$ for
every $x\in\mathcal{X}$. The supplier publishes an offer list
\begin{equation}
O_S(t)=\{(x,Q_x(t),P_{\mathrm{base}}(x))\}_{x\in\mathcal{X}},
\end{equation}
where $Q_x(t)$ is available supply and $P_{\mathrm{base}}(x)$ is a reserve (base) price.
We also generate a set of buyers $\mathcal{B}_t$ each step; each buyer $B_j$ has a latent persona text
$Persona_j$ and a buyer patience coefficient $\rho_j\in[0,1]$.
Importantly, personas are \emph{not} revealed to agents: retailers must infer demand preferences only through
public market outcomes, such as what sold and at what prices, which induces an incomplete-information setting.

\subsection{Stage A: Procurement as a Multi-Unit Auction}
Procurement implements a per-item, multi-unit, first-price auction with a reserve price. We deliberately choose first-price over second-price auctions because the latter reduces optimal policy to truthful bidding, whereas first-price auctions require bid shading, explicit competition modeling, and profit--win-rate tradeoffs,precisely the strategic reasoning we aim to evaluate.
Each agent submits bids of the form 
\begin{equation}
b_{i,x}(t)=(q_{i,x}(t),p^{\mathrm{bid}}_{i,x}(t))
\end{equation}
where $q_{i,x}$ is the
requested quantity and $p^{\mathrm{bid}}_{i,x}$ is the bid price.
Agents face a hard budget constraint (invalid bids receive zero allocation):
\begin{equation}
\sum_{x\in\mathcal{X}} q_{i,x}(t)\,p^{\mathrm{bid}}_{i,x}(t)\ \le\ Funds_i(t).
\end{equation}
The supplier allocates the available quantity of each item to the highest bids above the reserve price and charges
winners their bid prices, updating inventories and funds.

\subsection{Stage B: Retail as Price Competition with Persona-Gated Attention}
After procurement, each agent chooses (i) a retail price $P_{i,x}(t)$ for any subset of items, and (ii) a
short slogan $Slogan_i(t)$. The downstream market then matches buyers to retailers via a two-stage choice model:
\emph{attention} (who is visible) followed by \emph{purchase} (who is cheapest).

\paragraph{Persona-gated attention (consideration gate).}
For each buyer $B_j$, we embed both the slogan and persona with an embedding function $\mathbf{E}(\cdot)$ and
compute cosine similarity
\begin{equation}
\begin{aligned}
&\text{Sim}(i,j) \\
&= \cos\big(
\mathbf{E}(Slogan_i(t)),
\mathbf{E}(Persona_j)
\big)
\end{aligned}
\end{equation}
Buyers are sampled from a mixture of persona ``tribes'', including thrift, ethics, hype, and quality, each of which determines a persona template and a slogan sensitivity $\lambda_j\ge 0$. Attention weights are defined as
\begin{equation}
w_{i,j}=\exp\big(\lambda_j\,Sim(i,j)/\tau\big),
\end{equation}
with temperature $\tau>0$. The buyer then samples a consideration set $V_j(t)\subseteq\mathcal{A}$ in proportion to $w_{i,j}$; the set size is controlled by the buyer's patience coefficient $\rho_j$ and an upper bound $K_{\max}$. Even a low-price seller cannot sell if it is not sampled into $V_j(t)$.

\begin{table*}[htbp]
\centering
\resizebox{\textwidth}{!}{%
\begin{tabular}{c | l | c c c c | c c c c | c}
\toprule
\multirow{2}{*}[-0.8em]{\makecell{\textbf{Mode}\\\textbf{Type}}} & \multicolumn{1}{c|}{\multirow{2}{*}[-0.8em]{\textbf{Model}}} & \multicolumn{4}{c|}{\textbf{Economics}} & \multicolumn{4}{c|}{\textbf{Operation}} & \textbf{Cognitive} \\
\cmidrule(lr){3-6} \cmidrule(lr){7-10} \cmidrule(lr){11-11}
 &  & \textbf{NPM}$\uparrow$ & \textbf{Pi}$\uparrow$ & \textbf{RAR}$\uparrow$ & \textbf{IEI}$\uparrow$ & \makecell{\textbf{Stockout}\\\textbf{Rate}}$\downarrow$ & \makecell{\textbf{Bid}\\\textbf{Eff.}}$\uparrow$ & \textbf{OSI}$\uparrow$ & \makecell{\textbf{Fill}\\\textbf{Rate}}$\uparrow$ & \textbf{MMS}$\uparrow$ \\
\midrule
\multirow{5}{*}{\textbf{\makecell{Closed\\Source\\Model}}} & Gemini 2.5 Pro~\cite{gemini2025} & \textcolor{blue}{0.167} & \textcolor{red}{36589} & \textcolor{red}{2.751} & 0.861 & \textcolor{blue}{0.765} & \textcolor{blue}{0.792} & 0.777 & \textcolor{blue}{0.221} & \textcolor{red}{0.691} \\
 & Gemini 2.5 Flash~\cite{gemini2025} & \textcolor{red}{0.190} & \textcolor{blue}{26104} & \textcolor{blue}{1.569} & 0.645 & 0.791 & \textcolor{red}{0.795} & 0.820 & 0.194 & 0.675 \\
 & O3~\cite{o3_2025} & 0.097 & 13368 & 1.016 & \textcolor{blue}{0.901} & \textcolor{red}{0.573} & 0.699 & \textcolor{red}{1.000} & \textcolor{red}{0.418} & \textcolor{blue}{0.690} \\
 & Sonnet 4.5~\cite{sonnet45_2025} & 0.158 & 10619 & 1.438 & 0.780 & 0.892 & 0.411 & 0.603 & 0.097 & 0.671 \\
 & GPT-4o~\cite{gpt4o_2024} & 0.117 & 7619 & 0.712 & 0.742 & 0.911 & 0.450 & 0.640 & 0.079 & 0.669 \\
\midrule
\multirow{15}{*}{\textbf{\makecell{Open\\Source\\Model}}} & Phi-4~\cite{phi4_2024} & 0.110 & 7565 & 0.822 & 0.867 & 0.898 & 0.386 & 0.636 & 0.091 & 0.677 \\
 & Qwen2.5 VL 72B~\cite{qwen25vl_2025} & 0.058 & 3402 & 0.738 & 0.789 & 0.942 & 0.318 & 0.707 & 0.050 & 0.673 \\
 & Llama 3.1 70B~\cite{llama31_2024} & 0.068 & 2444 & 0.628 & 0.875 & 0.959 & 0.195 & 0.579 & 0.035 & 0.673 \\
 & QwenLong L1 32B~\cite{qwenlong_2025} & 0.085 & 2242 & 0.317 & 0.561 & 0.976 & 0.115 & 0.550 & 0.020 & 0.681 \\
 & Qwen2.5 32B~\cite{qwen25_2024} & 0.053 & 1815 & 0.562 & 0.856 & 0.964 & 0.189 & 0.582 & 0.031 & 0.660 \\
 & Gemma 3 27B~\cite{gemma3_2025} & 0.041 & 1481 & 0.580 & 0.862 & 0.949 & 0.296 & 0.732 & 0.043 & 0.680 \\
 & Qwen2.5 VL 32B~\cite{qwen25vl_2025} & 0.069 & 1409 & 0.656 & 0.609 & 0.967 & 0.197 & 0.540 & 0.026 & 0.639 \\
 & ERNIE 4.5 300B~\cite{ernie45_2025} & 0.037 & 1360 & 0.614 & 0.849 & 0.942 & 0.418 & 0.741 & 0.050 & 0.685 \\
 & InternLM2.5 20B~\cite{internlm2_2024} & 0.031 & 1154 & 0.440 & 0.843 & 0.959 & 0.441 & 0.801 & 0.035 & 0.675 \\
 & InternLM3 8B~\cite{internlm2_2024} & 0.028 & 1022 & 0.318 & 0.845 & 0.963 & 0.280 & 0.659 & 0.031 & 0.660 \\
 & DeepSeek V3.2~\cite{deepseekv32_2025} & 0.072 & 616 & 0.292 & 0.796 & 0.985 & 0.237 & \textcolor{blue}{0.829} & 0.013 & 0.685 \\
 & Qwen2.5 72B~\cite{qwen25_2024} & 0.021 & 335 & 0.309 & 0.759 & 0.973 & 0.140 & 0.554 & 0.022 & 0.674 \\
 & Hunyuan A13B~\cite{hunyuan_2024} & 0.014 & 298 & 0.134 & \textcolor{red}{0.941} & 0.975 & 0.152 & 0.591 & 0.020 & 0.683 \\
 & Qwen3 30B-A3B~\cite{qwen3_2025} & 0.000 & 0 & 0.000 & 0.000 & 1.000 & 0.000 & 0.796 & 0.000 & 0.677 \\
 & ERNIE 4.5 21B~\cite{ernie45_2025} & -0.004 & -58 & -0.039 & 0.547 & 0.985 & 0.216 & 0.571 & 0.015 & 0.659 \\
\bottomrule
\end{tabular}%
}
\caption{Performance Metrics of Different Models. For each metric, an arrow indicates whether higher ($\uparrow$) or lower ($\downarrow$) values are better. The best and second-best performances for each metric are highlighted in red and blue, respectively.}
\label{tab:mean-metrics}
\end{table*}

\begin{figure}[t]
\centering
\includegraphics[width=\columnwidth]{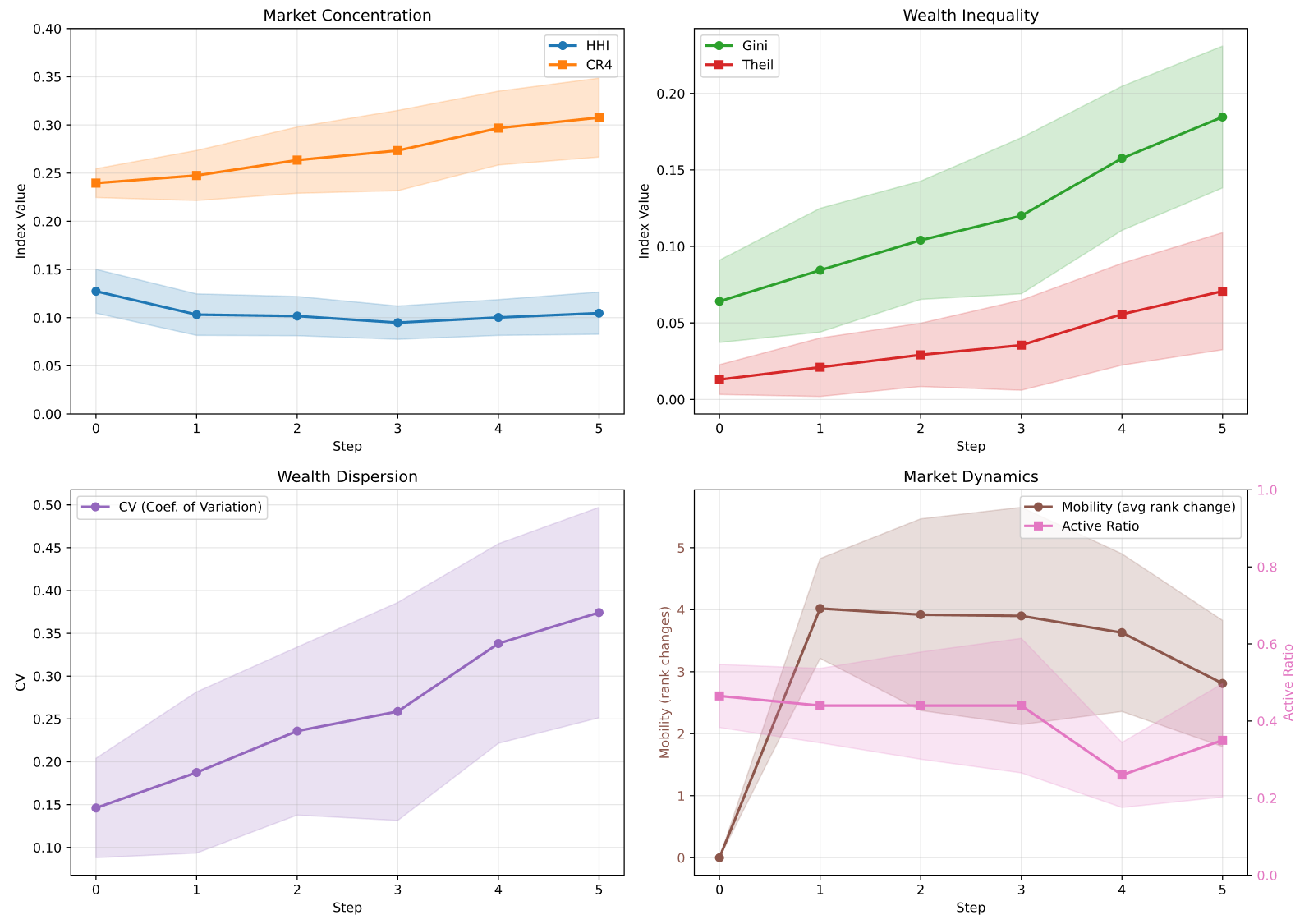}
\caption{Market-level indices computed from logged trajectories in the default setting. Inequality increases (Gini,
Theil, CV); the top-4 share rises (CR4) while concentration remains competitive (HHI); and the Active Ratio
declines.}
\label{fig:market-indices}
\end{figure}

\paragraph{Purchasing rule (price competition under constraints).}
At each step, buyers arrive with item demands under a controllable scarcity level (set by scaling aggregate demand relative to supply). For each desired item, the buyer considers only sellers in $V_j(t)$ who posted a price and have inventory, and purchases from the lowest-priced available seller until demand is met or inventory is exhausted. This creates Bertrand-style pressure on pricing within each buyer's attention set, while persona alignment controls which sellers compete for each buyer.

\subsection{Information Structure and Action Space}
At each step, agents observe their funds and inventory, the supplier offer list, and a compact public market history (posted prices, slogans, and realized sales) to support persona inference. They output structured numeric actions (item-level bids and retail prices) plus a free-form slogan, making Market-Bench a tightly coupled numeric-language decision problem.

\subsection{Agent Objective and Economic Trade-offs}
Market-Bench operationalizes standard firm objectives under scarce upstream supply. Let $y_{i,x}(t)$ be the number of units sold by agent $i$ for item $x$ at step $t$. Define procurement allocations $a_{i,x}(t)$ from Stage~A and retail revenues from Stage~B:
\begin{equation}
	    R_i(t)=\sum_x P_{i,x}(t)\,y_{i,x}(t)
\end{equation}
Agents aim to maximize cumulative profit and avoid bankruptcy by maintaining nonnegative funds. We report automatically computed metrics spanning economic outcomes, such as profit and net profit margin; operational outcomes, such as stockout rate and fill rate, plus bounded stability indices such as IEI and OSI; and language-persona alignment.

This structure induces coupled economic tensions: bidding aggressively secures upstream supply but increases unit cost and reduces future liquidity; pricing aggressively increases sell-through but erodes margin; and slogans must simultaneously differentiate the agent semantically (to enter more buyers' consideration sets) while remaining consistent with the agent's pricing and inventory strategy.

\subsection{Market-Bench as a Generative Dataset}
Unlike static test sets, Market-Bench produces complete interaction traces that can be treated as a dataset of strategic behavior. Each run logs, per step and per agent, the procurement bids and allocations, posted prices and slogans, realized sales events, and resulting balance-sheet state (funds and inventory). This makes it possible to evaluate not only \emph{final outcomes} but also \emph{process-level} properties such as how agents revise bids across rounds, how language changes with observed market feedback, and how numeric policies respond to inventory dynamics. Beyond agent-level scores, these logs enable \emph{dataset-level} analysis of market structure. We compute market-wide indices of inequality (Gini, Theil, CV), concentration (HHI, CR4), and participation (Active Ratio) directly from transaction outcomes. Figure~\ref{fig:market-indices} illustrates a typical evolution in our default setting: inequality rises quickly and participation declines, while the top-4 share increases even as HHI remains in the competitive range. These macro signals complement per-agent metrics and help characterize emergent dynamics in
Market-Bench.

\section{Experiments}
\label{sec:experiments}

\begin{figure*}[t]
\centering
\includegraphics[width=\textwidth]{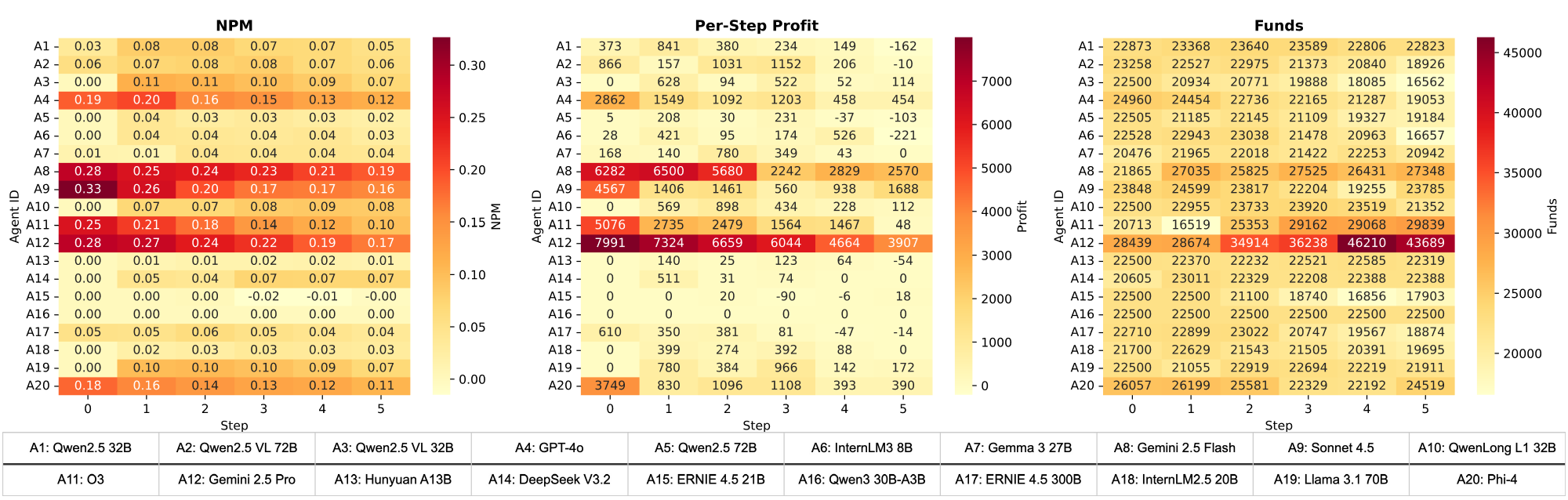}
\caption{Temporal economic outcomes (mean across 10 runs). From left to right: net profit margin (NPM), per-step profit,
and end-of-step funds. A small set of models, led by Gemini 2.5 Pro and Gemini 2.5 Flash, consistently earn high
profits and compound capital, while many models remain near break-even.}
\label{fig:econ-heatmap}
\end{figure*}

\subsection{Experimental Setup}
\paragraph{Simulator.}
We run the Market-Bench simulator (Section~\ref{sec:mercantilbench}) with $m{=}20$ retailer agents, each controlled by an LLM that outputs procurement bids, retail prices, and a marketing slogan. We evaluate the 20 LLM backends listed in Table~\ref{tab:mean-metrics} and log complete transaction-level trajectories for reproducibility.

\paragraph{Environment and initialization.}
The default setting uses 8 items with tiered base prices (50 / 150 / 800 / 2000) and quantities (200, 200, 133, 133, 134, 75, 75, 50), with a horizon of 6 steps and 2 bidding rounds per step.
Let total supplier supply be $S=\sum_x Q_x$ and total buyer demand be $D=rS$ with supply-demand ratio $r$ (we use $r{=}0.95$). Initial funds are set proportional to the supplier catalog value:
\begin{equation}
    K_{\mathrm{init}}=\alpha\cdot\frac{\sum_x Q_x P_{\mathrm{base}}(x)}{m}
\end{equation}
In our default large-scale setting,
\begin{equation}
    \sum_x Q_x P_{\mathrm{base}}(x)=300{,}000,\quad
    m{=}20,\quad \alpha{=}1.5,
\end{equation}
giving $K_{\mathrm{init}}{=}22{,}500$ per agent.
Buyer volume is configured as $k=\beta m$, with $k{=}200$ and $\beta{=}10$. Persona-Gated Attention uses $K_{\max}{=}20$ and $\tau{=}1.0$. All stress scenarios are disabled in these experiments to keep the economy stationary.

\subsection{Experimental Results}
We repeat the large-scale setting for 10 independent runs and report mean metrics across runs. Figures in this section report temporal economic, operational, and semantic dynamics aggregated over runs.

\begin{figure*}[t]
\centering
\includegraphics[width=\textwidth]{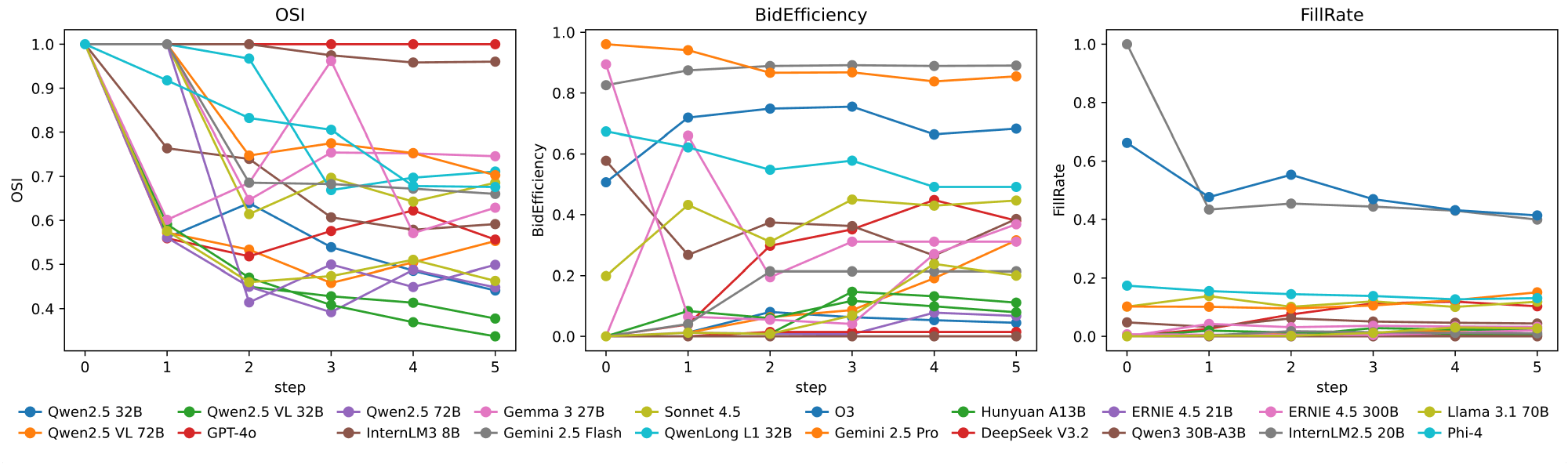}
\caption{Operational dynamics over time (mean across 10 runs). Left: Order Stability Index (OSI). Middle: procurement
BidEfficiency. Right: downstream FillRate. Agents with sustained auction success (high BidEfficiency) also achieve
higher FillRate and tend to maintain higher OSI.}
\label{fig:op-dynamics}
\end{figure*}

\begin{figure*}[t]
\centering
\includegraphics[width=\textwidth]{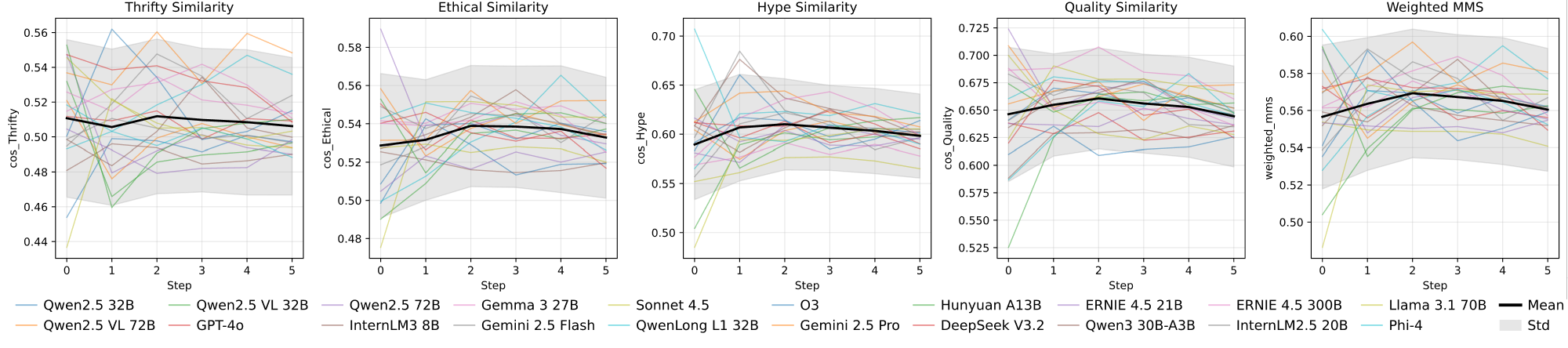}
\caption{Slogan-persona similarity dynamics (mean $\pm$ std across 10 runs). Each line is an agent's average cosine
similarity to each buyer tribe persona (Thrifty, Ethical, Hype, Quality) and the resulting weighted MMS. Across
agents, similarities shift most between steps 0 and 1 and then stabilize.}
\label{fig:tribe-similarities}
\end{figure*}

\begin{figure*}[t]
\centering
\includegraphics[width=\textwidth]{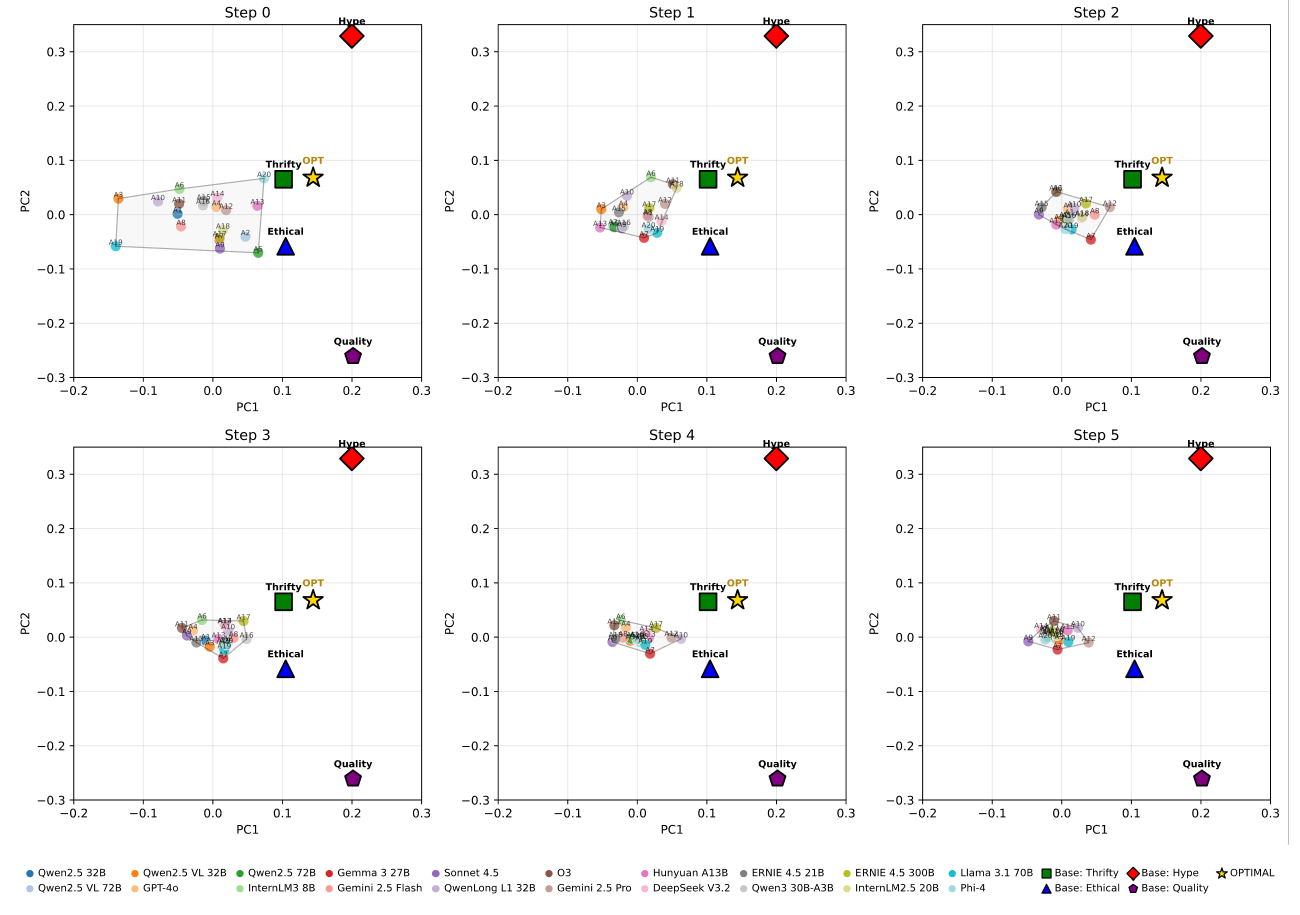}
\caption{Evolution of slogan embedding clusters across steps (PCA projection of tribe-similarity features; mean across
10 runs). Agents start dispersed but rapidly converge to a tight region by step 5, suggesting that competitive
feedback induces similar messaging strategies.}
\label{fig:slogan-cluster}
\end{figure*}

\subsection{Analysis}
Table~\ref{tab:mean-metrics} reveals performance dispersion that correlates with reasoning architecture rather than raw scale. Thinking-enabled models dominate: Gemini 2.5 Pro and Flash achieve the highest profits ($\Pi{=}36{,}589$ and $26{,}104$) and margins (NPM{=}0.167 and 0.190), while O3 attains the best service level (FillRate{=}0.418, OSI{=}1.000) through explicit chain-of-thought reasoning. Among open-source models, Phi-4 presents a striking result: with only 14B parameters, it achieves $\Pi{=}7{,}565$, comparable to GPT-4o ($\Pi{=}7{,}619$), suggesting that reasoning-focused training transfers effectively to economic tasks. In contrast, larger MoE models show mixed outcomes: DeepSeek V3.2 (671B total) achieves high operational stability (OSI{=}0.829) but modest profits, while Hunyuan A13B attains the highest inventory efficiency (IEI{=}0.941) yet fails to convert this into profitability. Notably, model scale alone is a poor predictor: Phi-4 (14B) outperforms all open-source models including ERNIE 4.5 300B ($\Pi{=}1{,}360$), while Qwen3 30B-A3B achieves zero profit with zero BidEfficiency across all runs, indicating complete failure to produce valid bids. This mode of failure, where a single formatting error leads to total market exclusion, is unique to economic benchmarks with hard constraints. Beyond these aggregate patterns, Figure~\ref{fig:econ-heatmap} reveals persistent stratification: a small set of models compounds capital early and maintains advantage, while others remain near break-even or drift into losses.

\paragraph{A rapid ``entry shock'' followed by stabilization.}
Across economic, operational, and semantic traces, the largest adjustments occur between steps 0 and 1. This is partly structural: OSI defaults to 1.0 at step 0 (insufficient points for variability), and then drops once order/sales variability becomes defined in step 1 (Figure~\ref{fig:op-dynamics}). Semantically, slogan-persona similarities shift most in the first step and then stabilize (Figure~\ref{fig:tribe-similarities}), while the embedding clusters collapse quickly and remain tight by step 5 (Figure~\ref{fig:slogan-cluster}). Together, these patterns suggest that early market feedback rapidly compresses the strategy space: after first interaction, agents mostly refine within a settled regime rather than continually exploring.

\paragraph{Inequality rises without monopoly: a multi-winner oligopoly.}
Market indices (Figure~\ref{fig:market-indices}) show wealth inequality tripling (Gini $0.07 {\rightarrow} 0.21$; Theil $0.02 {\rightarrow} 0.10$; CV $0.13 {\rightarrow} 0.45$), yet concentration stays competitive (HHI $\approx 0.08$--$0.10$) while the top-4 share rises (CR4 $0.23 {\rightarrow} 0.33$). This indicates an emergent ``multi-winner'' structure where a leading tier captures more share without monopoly. Rank mobility remains high, implying that relative ordering is contestable even as inequality grows, while the active ratio declines toward $\approx 0.4$ as many agents become effectively inactive.

\paragraph{Scarcity makes procurement a threshold skill, producing bimodality.}
Profit correlates strongly with procurement success: Spearman $\rho{=}0.68$ with BidEfficiency, $\rho{=}0.88$ with FillRate, and $\rho{=}{-}0.88$ with StockoutRate. Under scarcity, failure to secure inventory is a first-order error that pricing and language cannot compensate for. Figure~\ref{fig:op-dynamics} reveals a \emph{bimodal} regime: some agents sustain high BidEfficiency and meaningfully serve demand, while others remain near-zero. This separation is self-reinforcing,early success generates liquidity for future bids, whereas early failure produces revenue starvation,explaining the winner-take-most pattern in Figure~\ref{fig:econ-heatmap}.

\paragraph{Margin--volume trade-offs separate pricing from procurement.}
High FillRate does not guarantee high profit: o3 achieves the best FillRate but earns a lower margin than Gemini 2.5 Pro and Flash, suggesting aggressive bidding can trade margin for volume. Conversely, models with moderate service levels still profit through pricing discipline. Market-Bench thus distinguishes auction competence (acquiring supply) from retail competence (monetizing it). Figure~\ref{fig:econ-heatmap} shows margin compression over time, consistent with intensified competition as early rents dissipate.

\paragraph{Operational indices require deconfounding with activity.}
Bounded indices such as IEI and OSI help summarize process properties, but can be misleading without conditioning on participation. For example, a low-activity agent that rarely procures or sells may appear stable (high OSI) or efficient (high IEI) simply because there is little variability to measure. This motivates interpreting stability and efficiency jointly with market-access variables (BidEfficiency, FillRate, StockoutRate) and participation signals, rather than as standalone indicators of competence.

\paragraph{Language adapts quickly but converges toward a generic slogan optimum.}
MMS varies in a narrow band and is weakly correlated with profit (Spearman $\rho{=}0.16$). Figures~\ref{fig:tribe-similarities} and~\ref{fig:slogan-cluster} show that most slogan movement happens early and then converges, consistent with an emergent ``messaging equilibrium.'' Since the buyer distribution places substantial slogan-sensitive mass on the Ethical and Hype tribes, agents converge near the Ethical-Hype axis but do not separate into tribe-specialized niches, suggesting slogans are learned through mimicry rather than buyer analysis. We view this rapid convergence not as a benchmark limitation but as an empirical finding about current LLM agents: under competitive pressure they collapse to a safe mimicry equilibrium, failing to sustain defensible niche positioning. Breaking this equilibrium,and maintaining semantic differentiation under competition,is a concrete open challenge that Market-Bench enables future work to study.

\section{Conclusion}
\label{sec:conclusion}

Market-Bench provides a reproducible market simulation for evaluating LLM agents where numeric decisions and language jointly determine economic outcomes. By coupling budget-constrained procurement with downstream price competition and semantic visibility, the benchmark produces logged trajectories that serve as a dataset of strategic behaviors. Our experiments reveal a stark ``winner-take-most'' dynamic. We observe a bimodal distribution of outcomes where a small elite of agents consistently compounds capital, while the majority struggle to break even. This suggests that while models may excel at isolated tasks, the synthesis of inventory management with market positioning remains a distinct frontier of difficulty. Market-Bench thus offers a necessary testbed for the next generation of agents designed to navigate the full complexity of the economic world.

\section*{Limitations}
\label{sec:limitations}

\paragraph{Simulation Scope.}
The experiments presented use a single market configuration with 20 agents, 6 time steps, and 8 item types. While Market-Bench's Hydra-based configuration system supports extensive customization, including agent count, episode length, supply-demand ratios, holding cost rates, and bidding round counts. We have not exhaustively explored all parameter combinations. Future work could systematically vary these parameters to characterize how LLM agent performance changes under different market conditions, such as highly competitive (supply $\ll$ demand) or relaxed (supply $\gg$ demand) scenarios.

\paragraph{Buyer Model.}
The buyer persona distribution in our experiments is synthetically generated with fixed weights (Thrifty 40\%, Ethical 30\%, Hype 20\%, Quality 10\%). Although the framework allows custom persona definitions and sensitivity parameters ($\lambda$, $\rho$), we did not explore the full space of buyer heterogeneity. Additionally, buyers currently make single-item purchases without memory of prior interactions. Extending the framework to support basket purchases, repeat customers, and buyer learning dynamics would provide richer evaluation scenarios for future investigation.

\paragraph{Future Directions.}
The modular architecture of Market-Bench enables several natural extensions: (1) dynamic supply with stochastic uncertainty to test adaptive procurement strategies; (2) multi-market scenarios with cross-market arbitrage opportunities; (3) longer episode horizons to study the emergence of complex trading strategies; and (4) non-English markets to evaluate multilingual LLM performance in economic contexts. These directions would further stress-test the economic reasoning and strategic planning capabilities of LLM agents beyond the current benchmark scope.

\section*{Acknowledgments}
This work was supported in part by the National Natural Science Foundation of China under Grants 62225112, 62401365, 62522116, 62271312, 62132006, and U24A20220; in part by the China Postdoctoral Science Foundation under Grants BX20250411 and 2025M773473; in part by the STCSM under Grant 22DZ2229005; and in part by the New Generation Artificial Intelligence-National Science and Technology Major Project (2025ZD0124104) in collaboration with the Shanghai Artificial Intelligence Laboratory.

\bibliography{references}

\appendix

\section{Simulation Algorithm}
\label{app:algorithm}

The complete simulation loop for Market-Bench proceeds as follows. Each episode consists of $T$ steps, with each step comprising a procurement stage (Stage A) and a retail stage (Stage B).

\paragraph{Simulation Loop.}
\begin{enumerate}
\item \textbf{Initialize}: Set $\text{Funds}_i \gets K_{\mathrm{init}}$, $\text{Inv}_{i,x} \gets 0$ for all agents $i \in \mathcal{A}$ and items $x \in \mathcal{X}$.
\item \textbf{For each step} $t = 0, \ldots, T-1$:
\begin{enumerate}
\item Prepare supplier offers $O_S(t)$.
\item \textbf{Stage A (Procurement)}:
\begin{itemize}
\item For each bidding round $r = 1, \ldots, R_{\max}$:
\item Build bidding state with previous round results.
\item Each agent $i$ submits bid $b_i$ via LLM call (in parallel).
\item Validate budget constraints; reject overspending bids.
\end{itemize}
\item Settle final bids: allocate items to highest bidders above reserve price.
\item Update $\text{Funds}_i$ and $\text{Inv}_{i,x}$ based on allocations.
\item \textbf{Stage B (Retail)}:
\begin{itemize}
\item Each agent $i$ outputs prices $P_i$ and slogan via LLM call (in parallel).
\end{itemize}
\item Generate buyers $\mathcal{B}_t$ with persona embeddings.
\item For each buyer $B_j$: compute attention weights, sample consideration set, purchase from lowest-priced available seller.
\item Apply holding costs and check for bankruptcy.
\item Record metrics and market history.
\end{enumerate}
\item \textbf{Return}: Logged trajectories and final metrics.
\end{enumerate}

\subsection{Bid Settlement Procedure}
For each item $x$, bids are sorted by price descending with random tie-breaking. The available quantity $Q_x$ is allocated greedily to highest bidders above the reserve price $P_{\text{base}}(x)$:
\begin{equation}
\begin{aligned}
a_{i,x} &= \min(q_{i,x}, \text{remaining}) \\
\text{if } &p^{\text{bid}}_{i,x} \geq P_{\text{base}}(x)
\end{aligned}
\end{equation}

\section{LLM Agent Prompts}
\label{app:prompts}

\subsection{Procurement Stage Prompt}
The system prompt instructs the LLM to act as a retail agent in a sealed-bid auction:

\begin{quote}
\small
\texttt{You are a retail agent participating in a multi-round sealed-bid auction for supplier inventory. Output ONLY valid, parseable JSON. No prose.}

\texttt{Output schema (strict):}
\begin{verbatim}
{
 "bids": {
  "<item_id>": {"qty":<int>,"price":<int>}
 }
}
\end{verbatim}

\texttt{Rules:}
\begin{enumerate}
\item \texttt{Only bid on items in the supplier offers.}
\item \texttt{qty and price must be non-negative integers.}
\item \texttt{Do NOT overspend: total spend must not exceed Funds.}
\item \texttt{Do NOT bid below base\_price.}
\item \texttt{JSON must be strict (no trailing commas).}
\end{enumerate}
\end{quote}

\paragraph{Example Procurement Input.}
\begin{quote}
\small
\texttt{Step: 0} \\
\texttt{Round: 1 of 2} \\
\texttt{Funds: 22500} \\
\texttt{Overspent on most recent bid: False} \\
\texttt{Supplier offers:} \\
\texttt{item\_id,qty,base\_price} \\
\texttt{item1,200,50} \\
\texttt{item2,200,50} \\
\texttt{item3,133,150} \\
\texttt{...} \\
\texttt{Current inventory: item1,0; item2,0; ...} \\
\texttt{Historical sales: (empty at step 0)}
\end{quote}

\paragraph{Example Procurement Output (Gemini 2.5 Pro).}
\begin{verbatim}
{
  "bids": {
    "item1": {"qty": 110, "price": 51},
    "item2": {"qty": 110, "price": 51},
    "item3": {"qty": 75, "price": 151}
  }
}
\end{verbatim}

\subsection{Retail Stage Prompt}
The system prompt for the retail stage:

\begin{quote}
\small
\texttt{You are setting retail prices and a marketing slogan for your current catalog. Output ONLY valid, parseable JSON.}

\texttt{Output schema (strict):}
\begin{verbatim}
{
  "prices": { "<item_id>": <int>, ... },
  "slogan": "<string>"
}
\end{verbatim}

\texttt{Rules:}
\begin{enumerate}
\item \texttt{You do NOT know buyer personas; infer from market\_history.}
\item \texttt{price must be a non-negative integer.}
\item \texttt{Slogan should resonate with inferred personas.}
\item \texttt{Keep slogan short ($\leq$ 25 words).}
\end{enumerate}
\end{quote}

\paragraph{Example Retail Input.}
\begin{quote}
\small
\texttt{Step: 0} \\
\texttt{Funds: 10000} \\
\texttt{Current inventory:} \\
\texttt{item4: 19, item5: 19, item6: 3, item7: 3, item8: 1} \\
\texttt{Historical sales: (empty at step 0)}
\end{quote}

\paragraph{Example Retail Output (GPT-4o).}
\begin{verbatim}
{
  "prices": {
    "item4": 200,
    "item5": 200,
    "item6": 1000,
    "item7": 1000,
    "item8": 2500
  },
  "slogan": "Luxury and value, redefined!"
}
\end{verbatim}

\paragraph{Example Retail Output (Claude Sonnet 4.5).}
\begin{verbatim}
{
  "prices": {
    "item2": 75,
    "item4": 225,
    "item6": 1200,
    "item7": 1200,
    "item8": 3000
  },
  "slogan": "Premium quality at prices
             that make sense."
}
\end{verbatim}

\section{Evaluation Metric Definitions}
\label{app:metrics}

Market-Bench computes 9 per-agent metrics spanning economic, operational, and cognitive dimensions. Table~\ref{tab:mean-metrics} presents results using these metrics.

\subsection{Economic Metrics}

\paragraph{Net Profit Margin (NPM).}
Measures profitability relative to revenue:
\begin{equation}
\text{NPM} = \frac{\sum_{t=0}^{T-1} \Pi_i(t)}{\sum_{t=0}^{T-1} R_i(t) + \epsilon}
\end{equation}
where $\Pi_i(t) = R_i(t) - \text{COGS}_i(t) - H_i(t)$ is per-step profit, $R_i(t)$ is revenue, $\text{COGS}_i(t)$ is cost of goods sold, and $H_i(t)$ is holding cost. Higher is better.

\paragraph{Cumulative Profit (Pi).}
Total profit across all steps:
\begin{equation}
\Pi_i = \sum_{t=0}^{T-1} \Pi_i(t)
\end{equation}
Higher is better.

\paragraph{Risk-Adjusted Return (RAR).}
Sharpe-like ratio of mean profit to profit volatility:
\begin{equation}
\text{RAR}_i = \frac{\mu(\Pi_i(t))}{\sigma(\Pi_i(t)) + \epsilon}
\end{equation}
where $\mu(\Pi_i(t))$ and $\sigma(\Pi_i(t))$ are the mean and standard deviation of per-step profits. Higher is better.

\paragraph{Inventory Efficiency Index (IEI).}
Measures the fraction of goods sold relative to total throughput:
\begin{equation}
\text{IEI}_i = \frac{U^{\text{sold}}_i}{U^{\text{sold}}_i + \bar{I}_i + \epsilon}
\end{equation}
where $U^{\text{sold}}_i$ is total units sold and $\bar{I}_i$ is average inventory in units. IEI $\in [0, 1]$, higher is better.

\subsection{Operational Metrics}

\paragraph{Stockout Rate.}
Fraction of directed purchase attempts that failed due to zero inventory:
\begin{equation}
\text{StockoutRate}_i = \frac{N^{\text{stockout}}_i}{N^{\text{attempts}}_i + \epsilon}
\end{equation}
Lower is better.

\paragraph{Bid Efficiency.}
Combined measure of procurement success and cost efficiency:
\begin{equation}
\text{BidEff}_i = \frac{Q^{\text{win}}_i}{Q^{\text{bid}}_i + \epsilon} \times \frac{V^{\text{base}}_i}{C^{\text{win}}_i + \epsilon}
\end{equation}
where $Q^{\text{win}}_i$ is won quantity, $Q^{\text{bid}}_i$ is bid quantity, $V^{\text{base}}_i$ is base value of won goods, and $C^{\text{win}}_i$ is actual spend. Higher is better.

\paragraph{Order Stability Index (OSI).}
Measures how well order variability matches sales variability:
\begin{equation}
\text{OSI}_i = \frac{1}{1 + |\text{CV}_{\text{orders}} - \text{CV}_{\text{sales}}|}
\end{equation}
where the coefficient of variation is:
\begin{equation}
\text{CV}_X = \frac{\sigma_X}{\mu_X + \epsilon}
\end{equation}
OSI $\in [0, 1]$, higher is better.

\paragraph{Fill Rate.}
Fraction of directed demand (in units) successfully fulfilled:
\begin{equation}
\text{FillRate}_i = \frac{U^{\text{sold}}_i}{U^{\text{demand}}_i + \epsilon}
\end{equation}
Higher is better.

\subsection{Cognitive/Semantic Metric}

\paragraph{Mean Match Score (MMS).}
Average cosine similarity between agent slogans and buyer personas:
\begin{equation}
\text{MMS}_i = \frac{1}{|\mathcal{B}^{\text{interact}}_i|} \sum_{j} \text{Sim}(i, j)
\end{equation}
where $\text{Sim}(i, j) = \cos(\mathbf{E}(\text{Slogan}_i(t)), \mathbf{E}(\text{Persona}_j))$. Higher is better.

\section{Metric Replacement Rationale}
\label{app:metric-replacement}

Two traditional supply chain metrics were replaced with bounded alternatives due to numerical instability.

\subsection{IEI}
The Inventory Turnover Ratio (ITR = COGS / Avg.\ Inventory) becomes unbounded when average inventory approaches zero. Observed values ranged from 0 to $5.4 \times 10^{12}$, making comparison meaningless.

The Inventory Efficiency Index reformulates the metric:
\begin{equation}
\text{IEI} = \frac{U_{\text{sold}}}{U_{\text{sold}} + \bar{I} + \epsilon} \in [0, 1]
\end{equation}

\subsection{OSI}
The Bullwhip ratio (Var(Orders) / Var(Sales)) explodes when sales variance is near zero. Observed values ranged from 0.27 to $7.3 \times 10^{12}$.

The Order Stability Index uses coefficient of variation:
\begin{equation}
\text{OSI} = \frac{1}{1 + |\text{CV}_{\text{orders}} - \text{CV}_{\text{sales}}|} \in [0, 1]
\end{equation}

\section{Persona-Gated Attention Mechanism}
\label{app:attention}

The attention weight for buyer $j$ considering agent $i$ is:
\begin{equation}
\label{eq:attention}
w_{i,j} = \exp\left(\frac{\lambda_j \cdot \text{Sim}(i, j)}{\tau}\right)
\end{equation}
where $\text{Sim}(i, j) = \cos(\mathbf{E}(\text{Slogan}_i(t)), \mathbf{E}(\text{Persona}_j))$, $\lambda_j$ is slogan sensitivity, $\tau$ is semantic temperature, and $\mathbf{E}(\cdot)$ is the embedding function (Qwen3-Embedding-8B).

The consideration set size is:
\begin{equation}
|V_j| = \min\left(|\mathcal{A}|, \max(1, \lceil \rho_j \cdot K_{\max} \rceil)\right)
\end{equation}

\section{Buyer Persona Configuration}
\label{app:personas}

Table~\ref{tab:personas} shows the default buyer persona distribution.

\begin{table}[ht]
\centering
\small
\begin{tabular}{lccl}
\toprule
\textbf{Tribe} & \textbf{Weight} & $\lambda$ & \textbf{Keywords} \\
\midrule
Thrifty & 0.4 & 0.2 & (price-focused) \\
Ethical & 0.3 & 0.8 & green, fair, eco \\
Hype & 0.2 & 0.9 & exclusive, limited \\
Quality & 0.1 & 0.5 & quality, craft \\
\bottomrule
\end{tabular}
\caption{Default buyer persona distribution. $\lambda$ denotes slogan sensitivity weight.}
\label{tab:personas}
\end{table}

\section{Experimental Configuration}
\label{app:config}

Table~\ref{tab:config} summarizes the key simulation parameters.

\begin{table}[ht]
\centering
\small
\begin{tabular}{ll}
\toprule
\textbf{Parameter} & \textbf{Value} \\
\midrule
Number of agents & 20 \\
Number of steps & 6 \\
Bidding rounds & 2 \\
Buyers per round & 200 \\
Initial funds & 22,500 \\
\midrule
Item categories & 8 \\
Total supply units & 1,000 \\
Base prices & 50--2,000 \\
\midrule
Supply-demand ratio & 0.95 \\
Holding cost rate & 0.0 \\
Buyer patience ($\rho$) & 0.6 \\
\midrule
Embedding model & Qwen3-Embedding-8B \\
Temperature ($\tau$) & 1.0 \\
$K_{\max}$ & 20 \\
\midrule
LLM Temperature & 0.0 \\
Response format & JSON \\
\bottomrule
\end{tabular}
\caption{Default experimental configuration.}
\label{tab:config}
\end{table}

\section{Supplier Offer Structure}
\label{app:offers}

Table~\ref{tab:offers} details the supplier offer configuration.

\begin{table}[ht]
\centering
\small
\begin{tabular}{llrr}
\toprule
\textbf{Item} & \textbf{Category} & \textbf{Qty} & \textbf{Base Price} \\
\midrule
item1 & Commodity & 200 & 50 \\
item2 & Commodity & 200 & 50 \\
item3 & Standard & 133 & 150 \\
item4 & Standard & 133 & 150 \\
item5 & Standard & 134 & 150 \\
item6 & Luxury & 75 & 800 \\
item7 & Luxury & 75 & 800 \\
item8 & Veblen & 50 & 2,000 \\
\midrule
\multicolumn{2}{l}{Total} & 1,000 & -- \\
\bottomrule
\end{tabular}
\caption{Supplier offer structure for 20-agent experiments.}
\label{tab:offers}
\end{table}

\section{Evaluated Models}
\label{app:models}

Table~\ref{tab:models} lists the 20 LLM agents evaluated, spanning both open-weight and closed-source models.

\begin{table}[ht]
\centering
\small
\begin{tabular}{p{2cm}p{4.8cm}}
\toprule
\textbf{Category} & \textbf{Models} \\
\midrule
Closed-source & Gemini 2.5 Pro, Gemini 2.5 Flash, O3, Claude Sonnet 4.5, GPT-4o \\
\midrule
Open-weight & Phi-4, Qwen2.5 VL 72B, Llama 3.1 70B, QwenLong L1 32B, Qwen2.5 32B, Gemma 3 27B, Qwen2.5 VL 32B, ERNIE 4.5 300B, InternLM2.5 20B, InternLM3 8B, DeepSeek V3.2, Qwen2.5 72B, Hunyuan A13B, Qwen3 30B-A3B, ERNIE 4.5 21B \\
\bottomrule
\end{tabular}
\caption{LLM agents evaluated in Market-Bench experiments.}
\label{tab:models}
\end{table}

\end{document}